\documentclass{article}
\usepackage{spconf,amsmath,graphicx}
\usepackage{cite}
\usepackage{amssymb,amsfonts}
\usepackage{algorithmic}
\usepackage{graphicx}
\usepackage{textcomp}
\usepackage{xcolor}
\usepackage{enumitem}
\usepackage{lipsum}
\usepackage{caption}
\usepackage{subcaption}
\usepackage{enumitem}   
\usepackage[hidelinks]{hyperref}
\def\BibTeX{{\rm B\kern-.05em{\sc i\kern-.025em b}\kern-.08em
    T\kern-.1667em\lower.7ex\hbox{E}\kern-.125emX}}

\newcommand{\bm}{{\boldsymbol{\mathrm{m}}}}

\usepackage{tikz}
\usepackage{fancyhdr}
\definecolor{lime}{HTML}{A6CE39}
\DeclareRobustCommand{\orcidicon}{
    \begin{tikzpicture}
    \draw[lime, fill=lime] (0,0) 
    circle [radius=0.16] 
    node[white] {{\fontfamily{qag}\selectfont \tiny ID}};
    \draw[white, fill=white] (-0.0625,0.095) 
    circle [radius=0.007];
    \end{tikzpicture}
    \hspace{-2mm}
}

\foreach \x in {A, ..., Z}{\expandafter\xdef\csname orcid\x\endcsname{\noexpand\href{https://orcid.org/\csname orcidauthor\x\endcsname}
            {\noexpand\orcidicon}}
}

\title{Learned layered coding for Successive Refinement\\in the Wyner-Ziv Problem}

\name{Boris Joukovsky$^{1,2}$\orcidB{}\qquad Brent De Weerdt$^{1,2}$\orcidC{}\qquad Nikos Deligiannis$^{1,2}$\orcidA{}}

\address{$^1$ETRO Department, Vrije Universiteit Brussel (VUB), Pleinlaan 2, B-1050 Brussels, Belgium \\
$^2$imec, Kapeldreef 75, B-3001 Leuven, Belgium \\
}

\begin{document}


\maketitle

\ninept

\begin{abstract}
We propose a data-driven approach to explicitly learn the progressive encoding of a continuous source, which is successively decoded with increasing levels of quality and with the aid of correlated side information. This setup refers to the successive refinement of the Wyner-Ziv coding problem. Assuming ideal Slepian-Wolf coding, our approach employs recurrent neural networks (RNNs) to learn layered encoders and decoders for the quadratic Gaussian case. The models are trained by minimizing a variational bound on the rate-distortion function of the successively refined Wyner-Ziv coding problem. We demonstrate that RNNs can explicitly retrieve layered binning solutions akin to scalable nested quantization. Moreover, the rate-distortion performance of the scheme is on par with the corresponding monolithic Wyner-Ziv coding approach and is close to the rate-distortion bound.
\end{abstract}


\section{Introduction}

Distributed source coding (DSC) considers the separate encoding and joint decoding of samples of correlated sources. In the asymmetric scenario where one discrete source $Y$ is encoded independently, the Slepian-Wolf (SW) theorem~\cite{SW1973} implies that a correlated source $X$ can be decoded losslessly using $Y$ as side information, while being compressed at a rate identical to the case where the two sources are encoded jointly. Notably, this result has been generalized to Wyner-Ziv Coding (WZC) in the case of jointly Gaussian sources with square distortion measure~\cite{WZ1976}, extending SW coding to lossy compression. \textit{Successive refinement} for WZC is defined as the progressive, quality scalable coding of the source with available side information at the decoder. Crucially, successive refinability is achieved when no rate loss occurs compared to monolithic (non-scalable) coding, for instance, in the case of the joint Gaussian source and non-degraded side information~\cite{steinberg2004successive}.

Designing practical codes for WZC remains a challenge, in part because the achievability of SW coding (SWC) relies on a non-constructive, random binning procedure. Practical solutions approaching the rate-distortion bound include the use of trellis codes~\cite{pradhan2003distributed} or nested lattice codes~\cite{zamir2002nested} for quantization (sphere covering) and channel codes such as low density parity-check (LDPC) codes for SWC (sphere packing). The side information is used to decode both the channel and source codes. The approaches~\cite{Cheng05, Deligiannis14} to design successive refinable codes employ nested scalar quantization (NSQ) followed by SW encoding and decoding of successive bitplanes of the quantized signal using LDPC codes. However, NSQ requires very high dimensions to reach optimality and decoding LDPC codes is computationally expensive due to the use of belief propagation on a factor graph.

More recently, neural compression has been a subject of research in deep learning. The study in~\cite{balle2016end} has shown that black-box deep variational auto-encoders can learn a compressed representation of images, by optimizing a rate-distortion trade-off. Following these ideas, resolution scalable compression of images has also been studied in the non-distributed case~\cite{toderici2017full}, as well as quality scalable approaches~\cite{royen2021masklayer}. Only a few works have recently attempted to design neural compression schemes for WZC, with applications in distributed image compression~\cite{whang2021neural,mital2023neural}, multi-view image coding~\cite{zhang2022ldmic}, and federated learning topics, that is, distributed compression of gradients~\cite{abrahamyan2021learned} and federated compression learning~\cite{lei2022federated}. 
Despite their success, these existing neural WZC methods use heuristic approaches, where an autoencoder is used for separate encoding and joint decoding of correlated signals; these approaches are therefore not interpretable in the information theoretic sense of WZC.
Recent progress towards understanding learned WZ codes has been made in~\cite{ozyilkan2023learned}, which showed that artificial neural networks (ANNs) trained to minimize an upper bound on mutual information can explicitly learn random binning structures, under the assuption of ideal SWC or ideal entropy coding.  This is a first-of-its-kind result in data-driven approaches because it is akin to the achievability argument of the WZ theorem for jointly Gaussian sources.

In this paper, we present the first attempt to learn successive refinement for the Wyner-Ziv problem by extending~\cite{ozyilkan2023learned} to successively refined random binning. Specifically, we show that recurrent neural networks (RNNs) are suitable candidates to efficiently learn refinable encoders and decoders, akin to achieving layered WZC. Unlike vector quantized-based architectures~\cite{whang2021neural}, we rather let the model learn explicit quantizers that exhibit a behavior akin to binning of NSQ planes, which is known to be an asymptotically optimal solution to successively refinable WZC~\cite{Cheng05}. Our approach is based on optimizing two variational bounds on the asymptotic rates of each encoder, which are the rates of an ideal entropy coder and an ideal SW encoder, respectively. We show that the second case approaches the theoretical lower bound of layered WZC, while achieving equal performance to the monolithic scheme of~\cite{ozyilkan2023learned} and retrieving similar solutions to bit-plane binning.

The paper is organized as follows: Section~\ref{sec:background} describes the background on the successive refinement of the WZ Problem. Section~\ref{sec:method} derives the variational loss functions used to train the refinement stages and the proposed RNN-based architecture. Section~\ref{sec:results} presents the results obtained by training the model on correlated sources---whose correlation is modeled by an additive i.i.d. Gaussian noise---and explicitly displays the learned binning structure. Lastly, Section~\ref{sec:conclusion} concludes the work.


\section{Background}\label{sec:background}

\subsection{Wyner-Ziv coding}

We start by reminding the Wyner-Ziv theorem for DSC~\cite{WZ1976}. Consider a discrete, memoryless source $X \in \mathcal{X}^n$ and correlated side information $Y \in \mathcal{Y}^n$ known at the decoder-side only. The decoded signal $\hat{X}\in\hat{\mathcal{X}}^n$ is reconstructed with an average distortion $D$ based on the distance function $d(X,\hat{X})$, using $Y$ and a compressed bit string at some rate $R$ obtained by separate encoding of~$X$. The rate-distortion function $R_\mathrm{WZ}(D)$ defining the set of achievable rate-distortion pairs in this setup is given by:
\begin{equation}
\label{eq:WZrate}
    R_\mathrm{WZ}(D) = \min_{\substack{p(u|x) \\ \mathbb{E}d(X,\hat{X}) \leq D}} I(X;U|Y),
\end{equation}
where $U$ is an auxiliary variable satisfying the Markov chains $U \leftrightarrow X \leftrightarrow Y$ and $X \leftrightarrow (U, Y) \leftrightarrow \hat{X}$.
Remarkably, under certain conditions~\cite{WZ1976,pradhan2003duality,deligiannis2014no}, $R_\mathrm{WZ}(D)$ achieves the same rate-distortion performance as when the side information is known also at the encoder. 

\subsection{Successively refined Wyner-Ziv coding}

A successive refinement code with $K$~stages and non-degraded side information is defined as a set of encoder-decoder pairs $(\phi_k, \psi_k)$, where the decoders successively estimate $\hat{X}_k$ across distortion levels $(D_1, \dots, D_K)$ with the aid of the side information and previously accumulated codes:
\begin{align}
    &\phi_k: \mathcal{X}^n \rightarrow \{ 1, 2, \dots, M_k\},\\
    &\psi_k: \{ 1, \dots, M_1\} \times \dots \times \{ 1, \dots, M_k\} \times \mathcal{Y}^n \rightarrow \hat{\mathcal{X}}^n,
\end{align}
such that $\mathbb{E}d(X^n, \psi_k(\phi_1(X^n), \dots, \phi_k(X^n), Y^n)) \leq D_k$ and $M_k$ are alphabet sizes. The individual encoders communicate at rates $(R_1, \dots, R_K)$ and the total \textit{sum-rate} sent after $k$ stages is given by $R^\mathrm{tot}_{k} = \sum_{i=1}^k R_k$. The conditions for successive refinability in the Wyner-Ziv setting are stated next:

\textit{Theorem~\cite{steinberg2004successive}}: $X$ is said to be successively refinable with distortion levels $(D_1,D_2,\dots,D_K)$, if and only if there exists auxiliary random variables $U_k$ for $k=1,\dots,K$, with $K$ and the deterministic functions $f_k:\mathcal{U}_k\times\mathcal{Y} \rightarrow\hat{\mathcal{X}}$ such that the three following conditions hold, $\forall k:$
\begin{enumerate}[nosep]
    \item $R_\mathrm{WZ}(D_k) = I(X;U_k|Y)\ \mathrm{s.t.}\ \mathbb{E}[d(X, f_k(U_k, Y))] \leq D_k$,
    \item $(U_1,\dots,U_K) \leftrightarrow X \leftrightarrow Y$,
    \item $(U_1, \dots, U_{k-1}) \leftrightarrow (U_k, Y) \leftrightarrow X,\ k>1$,
\end{enumerate}
where $R_\mathrm{WZ}(D)$ is the Wyner-Ziv rate-distortion function. In other words, each stage must achieve the same optimal rate (in terms of sum-rate) as the monolithic (non successively coded) scheme. We may also note that the differential rate-distortion function, which characterizes the rate of each refinement layer, is given by:
\begin{equation}\label{diffrateWZ}
    R_\mathrm{WZ}(D_k) - R_\mathrm{WZ}(D_{k-1}) = I(X; U_k|U_{k-1},\dots,U_{1},Y)
\end{equation}
The study in~\cite{steinberg2004successive} proved that WZC is successively refinable when $X$ and $Y$ are joint Gaussian and the $\ell_2$ is the distortion metric. \cite{Cheng05} relaxed the condition to random sources correlated by additive \textit{i.i.d.} Gaussian noise.





\section{Learning Successively Refined Wyner-Ziv}\label{sec:method}
We aim to jointly learn the quantization, binning and reconstruction of a successively refinable source with side information at the decoder, with asymptotic rates close to the ideal WZ rate. In the way of~\cite{ozyilkan2023learned}, we seek to discover if a trained network can recover the NSQ and bit-plane binning solution, which is known to achieve successive refinability in WZC~\cite{Cheng05}. As for notations, the message resulting from the encoding of $x\sim X$ is denoted as $[m_1,\dots,m_K]$ where $m_k\in \{0,\dots,M-1\}$. The successive symbols $m_k$ are the outputs of $K$ encoders and are sent to $K$ corresponding decoders. We abbreviate $[m_1,\dots,m_k]$ with $m_1^k$ and likewise for sequences of any other variable. By abuse of notation, $m_1^k$ for $k\leq0$ refers to an empty sequence. Unless specified otherwise, $\log(\cdot)$ is the base-2 logarithm. In the following, we assume that the side information $Y$ is a noisy version of the source, such that $X=Y+N$, where $N\sim\mathcal{N}(0,\sigma_n^2)$ is Gaussian with different possible noise levels. This scenario is proven to be successively refinable in the WZ setting ~\cite{Cheng05}.

\subsection{Objective function}

We first discuss the training objective for a learned successive refinement model. For each refinement stage~$k$, with reconstruction outputs $\hat{x}_k$, we use as distortion metric
$\mathbb{E}\left[d(x, \hat{x}_k)\right]$,
where the expectation operator averages over all possible inputs~$x$ and reconstructions~$\hat{x}_k$, and the distance metric $d$ is the mean squared error (MSE). Recall then from Eq.~\eqref{diffrateWZ} that the differential rate of a single stage $R_k$ for the ideal WZ coder is given by:
\begin{align}
    R(D_k) &= \min I(X;U_k | U_1^{k-1},Y)  \\
    &= \min H(U_k | U_1^{k-1},Y) - H(U_k | U_1^{k-1}, X)  \\
    &= \mathbb{E}_{\substack{x,y\sim{}p(x,y) \\ u_1^k\sim p(u_1^k|x)}} \left[ \log\frac{p(u_k | u_1^{k-1},x)}{p(u_k | u_1^{k-1},y)} \right] \label{eq:diff_rate}
    ,
\end{align}
with the minimization constrained by $\mathbb{E}\left[d(x, \hat{x}_k)\right] \le D_k$.
When the coding problem is successively refinable, each stage~$k$ asymptotically achieves the WZ rate-distortion bound.

Since the distribution $p(u_k | u_1^{k-1},y)$ in general is not known in closed form, we define two upper bounds for \eqref{eq:diff_rate}, similar to~\cite{ozyilkan2023learned}:
\begin{align}
    R(D_k) &\le \mathbb{E}_{\substack{x,y\sim{}p(x,y) \\ u_1^k\sim p(u_1^k|x)}} \left[ \log\frac{p(u_k | u_1^{k-1},x)}{q_m(u_k | u_1^{k-1})} \right] ,  \label{eq:bound_marg}\\
    R(D_k) &\le \mathbb{E}_{\substack{x,y\sim{}p(x,y) \\ u_1^k\sim p(u_1^k|x)}} \left[ \log\frac{p(u_k | u_1^{k-1},x)}{q_c(u_k | u_1^{k-1},y)} \right] , \label{eq:bound_cond}
\end{align}
where $q_m$ and $q_c$ are two different models for the distribution $p(u_k | u_1^{k-1},y)$ and referred to as \textit{prior models}. These two bounds can be interpreted as corresponding to two different coding systems. The bound in~\eqref{eq:bound_marg}, using the marginal distribution $q_m(u_k | u_1^{k-1})$, is the rate of a system where the output of each encoder stage $p(u_k | u_1^{k-1},y)$ is further compressed using an ideal entropy coder at the rate $\mathbb{E}_x\left[ \mathbb{E}_{u_1^k\sim{}p(u_1^k | x)} \left[ -\log q_m(u_k | u_1^{k-1}) \right] \right]$. Eq.~\eqref{eq:bound_cond} uses the conditional distribution $q_c(u_k | u_1^{k-1},y)$, where the output of the encoder is compressed using an ideal Slepian-Wolf coder, asymptotically reaching a rate of $\mathbb{E}_{x,y}\left[ \mathbb{E}_{u_1^k\sim{}p(u_1^k | x)} \left[ -\log q_c(u_k | u_1^{k-1},y) \right] \right]$.

To establish the connection with the practical refinement code, we consider that each sample from $U_k$ in Theorem~1 represents the discrete encoder messages $[m_1, \dots, m_k]$, such that $p(u_k) = p(m_1^k)$ and $p(u_k | u_1^{k-1}) = p(m_k | m_1^{k-1})$. We also relax the hard constraint $\mathbb{E}\left[ d(x, \hat{x_k}) \right] \le D_k$ in the WZ theorem to obtain one of the two following losses for a single stage, depending on the choice of the upper bounds between~\eqref{eq:bound_marg}~or~\eqref{eq:bound_cond}:
\begin{align}
    \mathcal{L}_{m,k} &= \mathbb{E} \left[ \log\frac{p(m_k | m_1^{k-1},x)}{q_m(m_k | m_1^{k-1})} + \lambda\,d\mathopen{}\left(x, g_k(m_1^k, y)\right) \right]\mathclose{} ,\label{eq:loss_marg} \\
    \mathcal{L}_{c,k} &= \mathbb{E} \left[ \log\frac{p(m_k | m_1^{k-1},x)}{q_c(m_k | m_1^{k-1},y)} + \lambda\,d\mathopen{}\left(x, g_k(m_1^k, y)\right) \right]\mathclose{} \label{eq:loss_cond},
\end{align}
where $g_k(\cdot)$ is a joint decoder at stage $k$ with a continuous output, while the encoders and the prior models will be learned using discrete probabilistic models with categorical outputs. The derived loss functions minimize the rate and distortion simultaneously, balanced by hyperparameter $\lambda$, and the expectation operator subscripts are the same as in \eqref{eq:bound_marg}-\eqref{eq:bound_cond}, but are left out for brevity.
In practice, the complete, $K$-stages model is trained globally and end-to-end by summing the losses for each stage into the total loss function $\mathcal{L}_m = \sum_k \mathcal{L}_{m,k}$ or $\mathcal{L}_c = \sum_k \mathcal{L}_{c,k}$. The expectations will be approximated by training the models using stochastic gradient descent on large batches, and by sampling data samples $x,y$ from the input distribution and $[m_1,\dots,m_K]$ using the encoding functions $p(m_k|m_1^k,x)$. In the following section, we discuss the models used to learn the encoder and prior distributions, as well as the decoding functions.

\subsection{Proposed RNN-based successive refinement model}\label{model}

\begin{figure}[t]
\centering
     \begin{subfigure}[t]{\linewidth}
             \centering
             \includegraphics[width=1.0\textwidth]{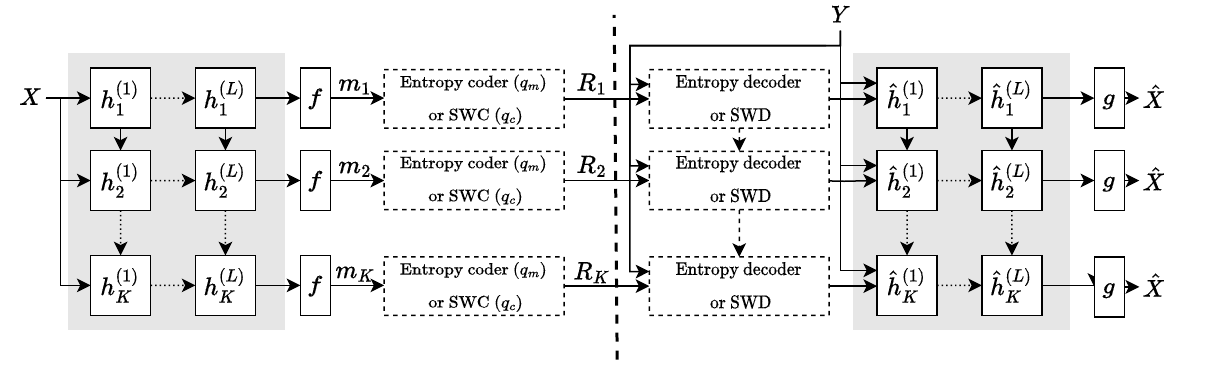}
             \caption{Proposed RNN-based encoders and decoders.}
             \label{fig:rnns}
     \end{subfigure}
    \begin{subfigure}[b]{\linewidth}
    \centering
    \vspace{0.3cm}
        \includegraphics[width=0.93\linewidth]{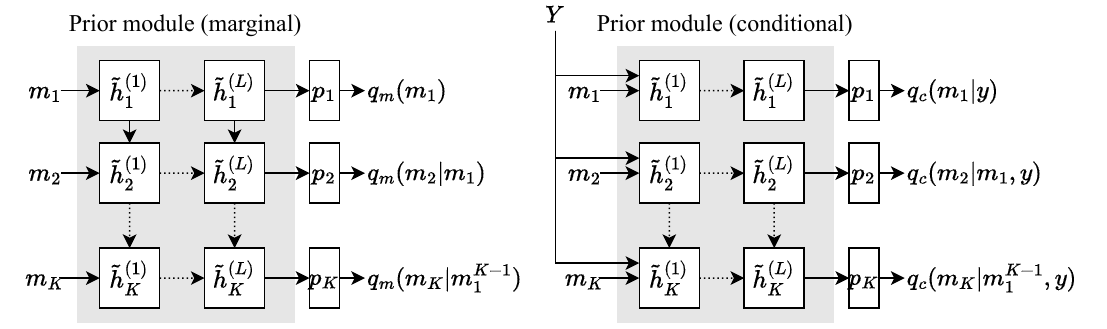}
        \caption{Marginal and conditional RNN-based prior modules.}
        \label{fig:priors}
    \end{subfigure}
    \vspace{-0.4cm}
    \caption{Our successive refinement model. Gray boxes are RNNs.}
    \label{fig:models}
    \vspace{-0.3cm}
\end{figure}
Our model is shown in Fig.~\ref{fig:models} and relies on stacked RNNs whose hidden states encode information of lower refinement levels. As depicted in Fig.~\ref{fig:rnns}, the $K$ encoders $p(m_k | m_1^{k-1},x)$ are parameterized by $\mathrm{softmax}(f(h_k^{(L)}))$, where $h_k^{(L)}$ is the hidden state of a stacked RNN with $L$ layers, taking the source realization $x$ as input for every time-step. We share a linear layer $f$ across all $K$ refinement stages. Therefore, after normalization, each encoder yields the probability vector $[p(m_k=0|m_1^{k-1},x), \dots, p(m_k=M| m_1^{k-1},x)]$, which is used in the previously derived loss functions. The successive decoders $g_k(m_1^k, y)$ are parameterized by $g(\hat{h}_k^{(L)})$, where $\hat{h}_k^{(L)}$ is the hidden state of another stacked RNN taking $y$ and the $k^{th}$ encoder output as input for every stage, assuming that the hidden state encodes the all the previously observed codes. The prior modules in Fig.~\ref{fig:priors} yield $q_m(m_k|m_1^{k-1})$ and $q_c(m_k|m_1^{k-1}, y)$ as $\mathrm{softmax}(p_k(\tilde{h}_k^{(L)}))$, where $p_k$ are linear mappings and $\tilde{h}_k^{(L)}$ is the output of yet another RNN taking at every stage $k$ either $m_{k-1}$ or the pair $(y, m_{k-1})$ as input, respectively. Setting $K=1$ reduces our model to the monolithic scheme in~\cite{ozyilkan2023learned}.

The motivation of using RNNs lies in the observation that the decision functions of a traditional NSQ---which is essential for layered WZC---are described by a quantization of successively finer fractional parts of the input and require modular arithmetics. In this context, we find that RNNs are suitable to learn a recursive function of the input that approximate the quantization of successive bit-planes of a NSQ, thereby allowing scalability at higher quality levels. Moreover, at the decoder, using a RNN with sequential inputs maintains a fixed input size (one-hot encoded bins) compared to using separate decoders with increasing complexity. Lastly, using a RNN diminishes the risk of training separate feed-forward NNs with a potential competition between the refinement stages.

During training, we use the same variational training approach as in~\cite{ozyilkan2023learned} by sampling the output of the encoder and the prior modules according to a Gumbel-softmax or Concrete distribution \cite{jang_categorical_2017, maddison_concrete_2017} with temperature $\tau$. This $\tau$ is decreased during training, thereby avoiding instability during the initial epochs and eventually making the samples $m_k$ approach a categorical (hard) distribution. During inference, the encoder is no longer probabilistic and the codes $m_k$ are set to $\mathrm{argmax}(f(h_k))$, encoded as one-hot vectors. This allows to estimate the theoretical rate for each level of the ideal entropy coder or SWC respectively, by summing over N samples $(x_n, y_n)$ drawn from $p(x,y)$:
\begin{align}
    R_{marg,k} &\approx \frac{1}{N}\sum_{n=1}^N-\log q_m(m_{n,k} | m_{n,1}^{k-1}) , \label{eq:rate_marg}\\
    R_{cond,k} &\approx \frac{1}{N}\sum_{n=1}^N-\log q_c(m_{n,k} | m_{n,1}^{k-1}, y_n) , \label{eq:rate_cond}
\end{align}
where the probabilities in Eq.~\eqref{eq:rate_marg}-\eqref{eq:rate_cond} are selected from the prior modules according to the predicted codes.

\section{Results}\label{sec:results}

All RNNs use 2 hidden layers with 100 hidden units each and with LeakyReLU activations, and linear layers are added at the output of each RNN, which are (Gumbel-)softmax-activated for the probabilistic models. We also compare with the monolithic scheme, by implementing~\cite{ozyilkan2023learned} and training it for different numbers of bins. All models, including the monolithic ones, are trained using the Adam optimizer with an initial learning rate of $10^{-3}$ for 180 epochs. The learning rate is decreased by a factor 0.3 every 80 epochs for the marginal distribution model in \eqref{eq:bound_marg}, and every 40 epochs for the conditional model in \eqref{eq:bound_cond}. The Gumbel-softmax temperature $\tau$ is decreased exponentially from 1.0 to 0.2 during training. We draw $2\times 10^{5}$ $(x, y)$ samples per epoch with a batch size of $10^3$, s.t. $Y\sim\mathcal{N}(0, 1)$ and $X = Y + N$ where $N\sim\mathcal{N}(0, \sigma_n^2)$. During testing, we estimate the rates and distortions over $10^{7}$ samples. For stability purposes, we use a stop-gradient operation on the predicted codes at the input of the prior modules as it was found to improve the convergence of the model. 

\begin{figure}
    \centering
    \includegraphics[width=\linewidth]{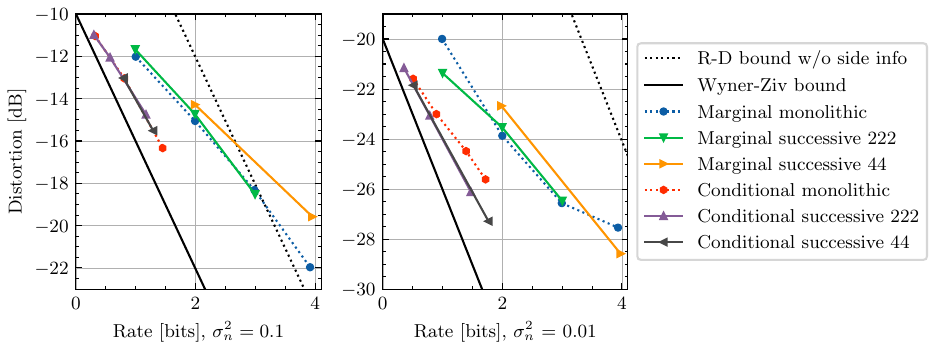}
    \caption{Rate-distortion curves for the monolithic model and the proposed successive refinement model in the 222 and 44 setting, for both the marginal and conditional rate estimations.}
    \label{fig:RDresults}
    \vspace{-0.3cm}
\end{figure}

We consider two successive refinement scenarios, and consequently two versions of the RNN models. In one case, we use three refinement stages, where the output of the encoder at each level is binary ($M=2$). This scenario is referred to as “222" The rate-distortion should thus be similar to the output of the monolithic model with 2, 4, and 8 bins respectively. The other scenario has two refinement stages with an alphabet size of $M=4$, called the “44" model. These two stages should then match the performance of the monolithic model with 4 and 16 bins, respectively. Furthermore, both marginal and conditional distribution models resulting from the upper bounds \eqref{eq:bound_marg}-\eqref{eq:bound_cond} are tested in both scenarios. The rate-distortion curves for all successive refinement RNNs are shown in Fig.~\ref{fig:RDresults}, along with the marginal and conditional monolithic models from \cite{ozyilkan2023learned}, the Wyner-Ziv bound given by~\eqref{eq:WZrate} and the rate-distortion bound for when no side information is available. We will discuss the results for the marginal and conditional models separately in the following sections.

\begin{figure}[t]
    \begin{subfigure}[t]{\linewidth}
        \centering
        \includegraphics[width=0.95\linewidth]{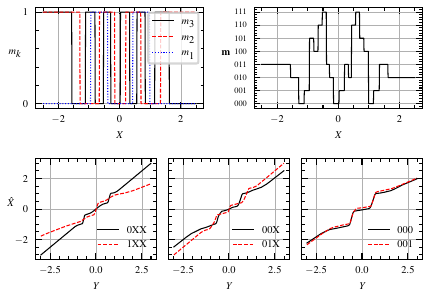}
        \vspace{-0.2cm}
        \caption{Conditional successive 222, $\sigma_n^2=0.01$.}
        \label{fig:binning222}
    \end{subfigure}
    \begin{subfigure}[b]{\linewidth}
        \centering
        \vspace{0.2cm}
        \includegraphics[width=0.95\linewidth]{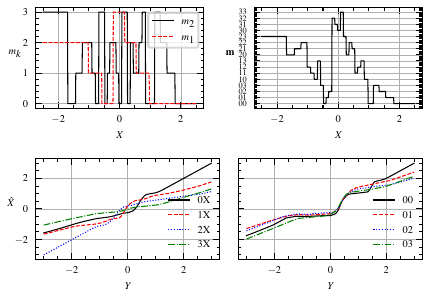}
        \vspace{-0.2cm}
        \caption{Conditional successive 44, $\sigma_n^2=0.01$.}
        \label{fig:binning44}
    \end{subfigure}
    \vspace{-0.4cm}
    \caption{Top-left: visualization of the learned quantizer or bins at each encoding layer. Top-right: complete message $\bm$, by concatenation of the encoders outputs. Bottom: successively refined input reconstruction w.r.t the side information and sent codes (e.g.: '0XX' shows the output of the first decoding stage with $m_1=0$.)}
    \label{fig:binning}
    \vspace{-0.3cm}
\end{figure}

\subsection{Marginal models}

As mentioned before, in this setup resulting from the bound~\eqref{eq:bound_marg}, the encoder needs to learn both the quantization of the signal and the entropy coding of the resulting bins. The rate of the system is calculated using the cross-entropy approximation of Eq.~\eqref{eq:rate_marg}. The same binning behavior is observed in both the marginal and conditional models, and illustrated for the conditional case in Fig.~\ref{fig:binning}.

For all models, we observe that the output distribution at each stage or number of bins is nearly uniform, resulting in rates only very slightly below $\log N$, with $N$ the number of bins.
In the left plot of Fig.~\ref{fig:RDresults}, we can see that the successive 222 coder performs nearly as well as the monolithic model for the noise level $\sigma_n^2=0.1$, with differences in distortion below 0.4~dB. For $\sigma_n^2=0.01$ in the right plot of Fig.~\ref{fig:RDresults}, the first stage of the successive 222 model improves over the monolithic model with a reduction of 1.4~dB in distortion.
The successive 44 model loses 0.7 and 2.4~dB in the case $\sigma_n^2=0.1$, while for $\sigma_n^2=0.01$ the model loses 1.3~dB at the first stage, but improves on the monolithic model at the second stage with 1.0~dB.

\subsection{Conditional models}

In this setup, resulting from the bound~\eqref{eq:bound_cond}, the encoder can be seen as a quantizer, while the distribution $q_c(m_k | m_1^{k-1},y)$ relates to the rate of an ideal SW coder and is estimated according to Eq.~\eqref{eq:rate_cond}. The results for this scenario are therefore closer to the WZ bound than the marginal models, since a larger part of the code is assumed to be ideal.
For $\sigma_n^2=0.1$ in Fig.~\ref{fig:RDresults}, the curves for the monolithic and successive refinement models are nearly identical, showing that we can achieve refinement without rate loss. For the low-noise case $\sigma_n^2=0.01$ in Fig.~\ref{fig:RDresults}, the layered coders even improve over the monolithic model, with a reduction in distortion of up to 1.7~dB.

Contrary to~\cite{ozyilkan2023learned}, we observe discontinuous binning of source samples, resembling the random binning in the Slepian-Wolf achievability theorem \cite{SW1973}, in both the marginal and conditional models. An illustration is given in Fig.~\ref{fig:binning222} for the conditional 222 model and in Fig.~\ref{fig:binning44} for the 44 model. In the top row, the output of the encoder is plotted for each stage, in function of the input sample $x$. The mappings learned by the RNNs are characterized by discontinuous intervals. Moreover, the decision boundaries appear to be interleaved between successive stages, and their frequency increases as $k$ increases. In the bottom row, we show the reconstruction function for some of the possible decoder inputs at each stage, in function of the side information. Moving towards the right-most plots highlights the successive refinement of the output for the selected bins. Similar to \cite{ozyilkan2023learned}, we observe nearly linear reconstruction functions within each bin, which is the theoretically optimal strategy.

\section{Conclusion}\label{sec:conclusion}

In this work, we designed the first learned successively refined model for Wyner-Ziv coding. We used two upper bounds on the Wyner-Ziv rate and added a distortion term to obtain two loss functions to train our RNN models. We showed that the learned successive models can attain a performance close to the state-of-the-art monolithic learned Wyner-Ziv coder, showing the viability of neural networks in successively  refined coding. Additionally, we showed that the models can learn nested quantization and binning behavior, similar to the operations in the achievability proof in the Wyner-Ziv theorem.

\bibliographystyle{IEEEbib}
\bibliography{biblio.bib}

\end{document}